\documentclass{article}

\usepackage{amssymb, amsmath, amsopn}
\usepackage{times,mathptmx}
\usepackage{mathptmx}
\usepackage{helvet}
\usepackage{courier}
\usepackage{makeidx}
\usepackage{multicol}
\usepackage{footmisc}

\usepackage{amssymb, amsmath}
\usepackage{graphicx}
\usepackage{caption}
\usepackage{subfig}
\usepackage{xspace}
\usepackage{color}
\usepackage{array}
\newcolumntype{P}[1]{>{\centering\arraybackslash}p{#1}}

\usepackage{verbatim}
\usepackage{imakeidx, epsfig,url}

\usepackage{xspace}
\usepackage[numbers,sort&compress,longnamesfirst,sectionbib]{natbib}
\usepackage{balance}

\usepackage{graphicx}

\begin{document}

	\title{A Feature-Based Comparison of Evolutionary Computing Techniques for Constrained Continuous Optimisation}
	\author{
		Shayan Poursoltan\\
		Optimisation and Logistics\\
		School of Computer Science\\
		The University of Adelaide\\
		Adelaide, Australia
		\and 
		Frank Neumann\\
		Optimisation and Logistics\\
		School of Computer Science\\
		The University of Adelaide\\
		Adelaide, Australia
	}

	\maketitle
\begin{abstract}
	Evolutionary algorithms have been frequently applied to constrained continuous optimisation problems.
	We carry out feature based comparisons of different types of evolutionary algorithms such as evolution strategies, differential evolution and particle swarm optimisation for constrained continuous optimisation. In our study, we examine how sets of constraints influence the difficulty of obtaining close to optimal solutions. Using a multi-objective approach, we evolve constrained continuous problems having a set of linear and/or quadratic constraints where the different evolutionary approaches show a significant difference in performance. Afterwards,  we discuss the features of the constraints that exhibit a difference in performance of the different evolutionary approaches under consideration.
	
\end{abstract}

\section{Introduction}
There have been many algorithmic approaches proposed to solve complex optimisation problems, including constrained optimisation problems (COP). Several approaches have been proposed to tackle the constraints in constrained problems. Most of the research has been focused on introducing differential evolution (DE) \cite{storn1997differential}, particle swarm optimisation (PSO) \cite{eberhart1995new} and evolutionary strategies (ES) \cite{schwefel1993evolution} to solve numerical optimisation problems. In order to deal with these constrained problems, there have been techniques that applied to these algorithms such as penalty functions, special operators (separating the constraint and objective function treatment) and decoder based methods. We refer the reader for a survey of constraint handling techniques in evolutionary computing methods to \cite{mezura2011constraint}.

In order to compare and evaluate the evolutionary algorithms many approaches have been used. One is finding which algorithm performs better on a set of continuous problems using benchmarks sets \cite{mallipeddi2010problem,hansen2010real}. Recently, there has been an increasing interest to analyse the problem features that make it hard to solve. Initial studies have been carried out in the field of continuous optimisation in \cite{mersmann2010benchmarking}. Furthermore, there have been techniques that generate a variation of problem instances from easy to hard. Then, the features of this problem instances are analysed in order to find which of them make the problems hard or easy to solve. Generating the variety of problem instances from easy to hard ensures that the knowledge obtained from analysis is reliable.

Although there is not only a standalone feature that makes a problem hard to solve, but it is assumed that constraints are very important in constrained continuous problems. The evolving approach that has been used to analyse the constraint features and their effects on COP's difficulty is discussed in \cite{poursoltan2014feature,shayan2015}. The idea is to evolve constrained problem instances (by using an evolutionary algorithm) in order to identify the constraint features with more contribution to problem difficulty. 

In this paper, by using a single-objective evolutionary algorithm, we generate hard and easy COP instances for DE, ES and PSO algorithms. Later, we solve the generated instances using one algorithm by the other algorithms. The results show that the hardest generated instances using one algorithm are still hard for the other ones. To get better insight, we use multi-objective evolving approach to generate instances that are hard for one algorithm but still easy for the others. By analysing how an algorithm fails in conditions where the rest perform well, we can derive its strengths and weaknesses over constraint features. Our study shows the effectiveness of constraint features that make the problems hard for one and easy for the other algorithms. It can be translated as over which features of constraints, they make the problems hard for a certain algorithm but still easy for the others.

The remainder of this paper is as follows: In Section 2 we introduce the concept of COPs. Then we discuss the evolver (single and multi-objective evolutionary approach) and the solver algorithms (DE, ES and PSO) we use in our experiments. In Section 3 we analyse the performance of various algorithms on each others hard and easy instances (using the single-objective evolver). Section 4 includes the multi-objective approach that generates hard instances for one but easy for the other algorithms. Furthermore, we carry out the analysis of linear and quadratic constraint features that make the problem hard for one and still easy for the rest. Finally, we conclude with some remarks.

\section{Preliminaries}
\subsection{Constrained continuous optimisation problems}
In this study, constrained continuous optimisation problems with inequality and equality constraints are investigated. These problems are optimisation problems where a function $f(x)$ should be optimised with respect to a given set of constraints.

Single-objective functions $f \colon S \rightarrow \mathbb{R}$ with $S \subseteq \mathbb{R}^n$ are considered in this research.
The constraints impose a feasible subset $F \subseteq S$ of the search space $S$ and the aim is finding $x \in S \cap F$ which minimises $f$. Formally, we state the problems as follows:

	\begin{equation}
		\begin{split}
		\text{minimize } &\quad    f(x), \quad	x = (x_1,\ldots,x_n) \in \mathbb{R}^n  \\
		\text{subject to}&\quad	g_i(x)  \leq 0 \quad  \forall   i \in \{1,\ldots,q\}\\
		&	\quad	h_j(x) = 0 \quad \forall j \in \{q + 1,\ldots, p\} 
		\label{eq:f}
		\end{split}
		\end{equation}

where $x = (x_1,x_2,\dots,x_n)$ is an $n$ dimensional vector and $x \in S  \cap F$. The $g_{i}(x)$ (inequality) and $h_{j}(x)$ (equality) constraints could be linear/nonlinear. Also,
the equality constraints are usually replaced by $|h_j(x)|  \leq \epsilon$ where $\epsilon=10e^{-4}$ \cite{mallipeddi2010problem}. The feasible region $F \subseteq S$ of the search space $S$ is defined by

\begin{equation}
l_i  \leq x_i  \leq u_i,\quad  \quad     1 \leq i  \leq n
\label{equality}
\end{equation}

\noindent where $l_i$  and $u_i$ denote lower and upper bounds respectively for the $i$th variable in which  $1\leq i \leq n$.
In this paper, we focus on the ability of constraints (linear, quadratic) to make a problem hard or easy. The features of these constraints and their effect on problem difficulty is discussed. The constraints are of the following form: 

\begin{equation}
\text{linear constraint} \quad	g(x) = b + a_{1}x_{1} + \ldots + a_{n}x_n
\label{lineareq}
\end{equation}

\begin{equation}
\text{quadratic constraint} \quad	g(x) = b +a_{1}x_{1}^{2} + a_{2}x_{1}\ldots + a_{2n-1}x_n^{2} +a_{2n}x_n
\label{quadraticeq}
\end{equation}

\noindent or a combination of them, where $x_{1},x_{2} \dots ,x_{n}$ are values from Equation 1 and $a_{1},a_{2},\dots,a_{n}$ are coefficients within lower ($l_{i}$) and upper bounds ($u_{i}$). We assume univariant quadratic function to analyse each $x_{n}$ (with exponent 2) independently. Also, unvivarient quadratic constraints are more popular in recent benchmarks \cite{mallipeddi2010problem}.
In order to include the optimum of objective function in feasible area, we set $b\leq0$ (we assume the objective function optimum is zero). 
\subsection{Algorithms}
We now introduce the algorithms for constrained continuous optimisation that are subject to our investigation. 

One of the most prominent evolutionary algorithms for COPs is $\epsilon$-constrained differential evolution with an archive and gradient-based mutation ($\epsilon$DEag). The algorithm is the winner of 2010 CEC competition for continuous COPs \cite{mallipeddi2010problem}. The $\epsilon$DEag uses $\epsilon$-constrained method to transform algorithms for unconstrained problems to constrained ones. It adopts $\epsilon$-level comparison to order the possible solutions. In other words, the lexicographic order is used in which constraint violation ($\phi(x)$) has more priority and proceeds the function value ($f(x)$). For more details we refer the reader to \cite{takahama2010constrained}.

The second algorithm we use in this paper is a $(1+1)$ CMA-ES for constrained optimisation \cite{arnold20121+}. The $(1+1)$ CMA-ES in \cite{igel2006computational} is a variant of  $(1+1)$-ES which adapts the covariance matrix of its offspring distribution in addition to its global step size. The idea behind the constraint handling approach of this algorithm is to obtain approximations to the normal vectors directions in the vicinity of the current solutions locations by low-pass filtering steps which violates the respective constraints and reducing the variance of the offspring distribution in these directions. Incorporating this constraint handling approach with $(1+1)$ CMA-ES makes an algorithm which is significantly more efficient than other approaches for constrained evolutionary algorithms. Also, the selected algorithm is not sensitive to the rotation of the problem search space. We refer the reader to \cite{arnold20121+} for more details and implementation.

The third algorithm that is used in our investigation is a particle swarm optimisation. This algorithm (HMPSO) applies a method that uses parallel search operator in which it divides the current swarm into various sub-swarms and locates the solution between them. In each sub-swarm, all particles follow the local best (fittest particle) which improves them to be more fitter. Also, since all sub-swarms are located around different optima (in parallel), then it is more possible to locate multiple optima which improves the diversity of algorithm. Dividing the swarms into sub-swarms improves the diversity of the algorithm. Also, choosing the local best in each sub-swarm can attract the other particles to fitter positions. We refer the reader to \cite{wang2009hybrid} for detailed algorithm and implementation.

\subsection{Features of Constraints}
In this paper we analyse the constraint features of generated problem instances. These features are constraint coefficients relationships such as standard derivation, angle between constraint hyperplanes, feasibility ratio in vicinity of optimum, number of constraints, shortest distance of constraint hyperplane to optimum. The details of these features are discussed in \cite{shayan2015}.

\section{Single-objective Investigations}

We first consider different algorithms and compare their relative performance on each other's generated hard and easy instances. We use single-objective evolver to evolve and generate hard and easy instances for all types of algorithms. The detailed procedure and results for DE instances are discussed in \cite{shayan2015}. For this experiment, we perform 30 independent runs generating easy and hard instances for PSO and ES solvers. It means, the single-objective evolver only generates instances that are hard/easy for one type of algorithm (PSO, ES and DE). The required function evaluation number (FEN) for solving these instances (PSO, ES and DE) is used as fitness value for single-objective evolver. The parameters for solvers are identical to \cite{takahama2010constrained,wang2009hybrid,arnold20121+}. Also, we run our experiments on Sphere function (bowl shaped)\cite{hansen2010real}. We now have three groups of easy and hard instances generated for DE, ES and PSO algorithms. We then compare the DE, ES and PSO algorithms by applying them on each other's easy and hard instances. The analysis is done by comparing the required FEN for an algorithm to solve the other's generated problem instances. Then, it is possible to derive strengths and weaknesses of the considered algorithms by observing how well one algorithm performs in conditions where the other algorithms fail (or it is difficult for them). 
Table \ref{table:perflinearsphere} and \ref{table:perfquadsphere} show different algorithms performance on Sphere objective functions with linear/quadratic constraints (1 to 5 constraints). We also run our experiments on different objective functions such as Ackley and Rosenbrock. The results are shown in Tables \ref{table:perflinearackley}, \ref{table:perfquadackley}, \ref{table:perflinearrosenbrock} and \ref{table:perfquadrosenbrock}. It is interesting that all objective functions follow similar pattern.  Considering the required FEN to solve each instances, it is observed that hard instances are still the hardest for their own algorithms and hard for the others. It implies that the hard instances share some common features to make it difficult to solve for all solvers. However, the obtained knowledge is not enough to compare the algorithm capabilities to solve hard problem instances.

  \section{Multi-objective Investigations}

  Based on the experiment results in previous section, hard instances for each algorithm are still hard for the others. In order to extract more useful knowledge about the strengths or weaknesses of certain algorithms on constraint algorithms, we need problem instances that are hard for one and easy for the others. Analysing the features of these instances helps us extracting knowledge regarding the strengths and weaknesses of algorithms by examining why an algorithm performs better on some groups of features while the others fails. This will help us developing more efficient prediction model for automated algorithm selection.

  To do this, we use a multi-objective DE algorithm (DEMO) described in \cite{robivc2005demo} to minimise the FEN for one algorithm and maximise it for the others. In other words, the FEN for generated problem instances is higher (harder) for a certain algorithm and lower (easier) for the others. In order to find instances that are hard for one algorithm type and easy for the others, we need to find solution as diverse as possible. Also, the solutions need to be close to pareto front. Satisfying these two aims makes us to use multi-objective evolutionary algorithm to generate problem instances. Hence, we use differential evolution for multi-objective optimisation (DEMO) proposed by Robic in \cite{robivc2005demo}. Based on results in \cite{robivc2005demo}, the DEMO achieves efficiently the above two goals. In DEMO, the candidate solution replaces parent when it dominates it and if the parent dominates it, the candidate is discarded. Otherwise, if the candidate and parent cannot dominate each other, the candidate is added to the population. The major difference between DEMO and other multi-objective evolutionary algorithms is that the newly generated good candidates are immediately used in creation of the subsequent candidates. This improves fast convergence to the true pareto front, while the use of non-dominated sorting and crowding distance metric in truncation of the extended population promotes the uniform
  spread of solutions. We refer the reader to \cite{robivc2005demo} for further details and implementation.
  
  In the following, we discuss the results for algorithms performances comparison. We carry out 30 independent runs for each number of constraints that are hard for one algorithm but still easy for the others. We set the evolving algorithm (DEMO) generation number to 5000 and the other parameters of evolving algorithm are set to pop size = $40$, CR = $0.5$, scaling factor = $0.9$ and $FEN_{max}$ is $300K$. Values for these parameters have been obtained by optimising the performance of the evolving algorithm in order to achieve the more easier and harder problem instances. For each of three algorithms, their best parameters are chosen \cite{takahama2010constrained,wang2009hybrid,hansen2010real}. First, the ($\epsilon$DEg) algorithm parameters are considered as: generation number = $1500$, pop size = $100$, CR = $0.5$, scaling factor = $0.5$. Also, the parameters for e-constraint method are described in \cite{shayan2015}. Moreover, for evolutionary strategy we perform $(1,7)$-ES algorithm with $1500$ generation using $P_{f} = 0.4$ with tendency to focus on feasible solution. In HMPSO algorithm, the swarm size $N$ is set to $60$, each sub-swarm size ($N_{s}$) is $8$ and all the PSO parameters are considered as Krohling and Coelho's PSO \cite{krohling2006coevolutionary}. In order to solve generated COPs, HMPSO generation number is set to $1500$. We need to say the parameters for the solvers are identical to those given in \cite{takahama2010constrained,wang2009hybrid,arnold20121+,robivc2005demo} 
  
  In our all experiments, we generate set of problem instances that are hard to one algorithm and easy to the other ones. Tables \ref{table:fenmultilinsphere}, \ref{table:fenmultiquadsphere}, \ref{table:fenmultilinAckley}, \ref{table:fenmultiquadAckley}, \ref{table:fenmultilinRosen} and \ref{table:fenmultiquadRosen} show the function evaluation number (FEN) required for each algorithm to solve DE/ES/PSO hard instances for Sphere, Ackley and Rosenbrock objective functions (with 1 to 5 linear/quadratic constraints). As it is observed, there is more difference between the required FEN of instances generated by multi-objective algorithm evolver than the single-objective one.
  For instance, the required FEN for solving DE hard instances are higher for DE algorithm than solving it by ES and PSO algorithm. It means the DE hard instances are only hard for DE algorithm and easy for the others. In the following we start analysing constraint features of instances that are hard for one and easy for others.
  
\begin{table}
	\hspace{-1.0cm}
	\resizebox{0.45\columnwidth}{!}{%
		\parbox{.6\linewidth}{
			\centering

			\caption {The comparison of algorithms performance on each other's easy and hard instances based on required FEN for \textbf{Sphere} objective function with \textbf{linear} constraints. DE Easy (1 c) means instances that are easy for DE and with 1 constraint.}
			{%
				\begin{tabular}{|p{2.2cm}|c|c|c|c|}
					
					\hline \hline
					Instances	& DE algorithm & ES algorithm & PSO algorithm   \\ \hline
					DE Easy (1 c) &  25.6K &   28.2K & 33.2K  \\ \hline
					ES Easy (1 c) & 26.3K  & 27.1K   & 33.9K  \\ \hline
					PSO Easy (1 c) &  24.9K & 29.1K  & 72.5K   \\ \hline
					DE Easy (2 c) & 28.9K  & 21.9K   & 32.1K  \\ \hline
					ES Easy (2 c) &  25.2K &  24.3K  & 29.4K  \\ \hline
					PSO Easy (2 c) & 24.2K  & 25.2K   & 33.5K  \\ \hline
					DE Easy (3 c) & 32.4K  & 31.2K   & 33.9K  \\ \hline
					ES Easy (3 c) & 31.8K  &  29.1K  & 33.2K  \\ \hline
					PSO Easy (3 c) & 35.1K  & 28.6K   & 35.1K  \\ \hline
					DE Easy (4 c) & 34.2K  & 29.8K   & 38.2K  \\ \hline
					ES Easy (4 c) & 32.1K  & 31.5K   & 36.1K  \\ \hline
					PSO Easy (4 c) & 35.7K  & 28.9K   & 39.5K  \\ \hline
					DE Easy (5 c) &  35.3K &  42.1K  & 46.4K  \\ \hline
					ES Easy (5 c) &  31.2K &  45.2K  & 38.2K  \\ \hline
					PSO Easy (5 c) & 35.3K  & 44.9K   & 41.2K  \\ \hline \hline
					DE Hard (1 c) & 91.2K  & 78.3K   & 76.4K  \\ \hline
					ES Hard (1 c) &  81.3K &  86.4K  & 78.8K  \\ \hline
					PSO Hard (1 c) & 82.5K  & 72.5K   & 85.4K  \\ \hline
					DE Hard (2 c) & 93.4K &  81.3K  & 81.4K  \\ \hline
					ES Hard (2 c) & 84.3K  & 92.6K   & 79.4K  \\ \hline
					PSO Hard (2 c) &  85.7K & 84.1K   & 89.4K  \\ \hline
					DE Hard (3 c) & 98.3K  &  93.8K  & 78.9K  \\ \hline
					ES Hard (3 c) & 91.4K  & 108.6K  & 81.2K  \\ \hline
					PSO Hard (3 c) & 89.1K  & 98.2K   & 91.6K  \\ \hline
					DE Hard (4 c) & 104.2K  & 89.4K   & 82.5K  \\ \hline
					ES Hard (4 c) & 89.4K  & 115.1K   &  78.4K \\ \hline
					PSO Hard (4 c) &  92.9K &  93.5K  &  115.3K \\ \hline
					DE Hard (5 c) &  123.2K & 111.4K   & 98.4K  \\ \hline
					ES Hard (5 c) & 98.2K  &  133.2K  & 94.9K  \\ \hline
					PSO Hard (5 c) &  101.3K & 109.2K & 118.3K   \\ \hline
				\end{tabular}}
				\setlength{\tabcolsep}{5em}
				
				\label{table:perflinearsphere}
				
			}
		}
		\hfill
		\resizebox{0.45\columnwidth}{!}{%
			\parbox{.6\linewidth}{
				\centering
				\caption {The comparison of algorithms performance on each other's easy and hard instances based on required FEN for \textbf{Sphere} objective function with \textbf{quadratic} constraints. DE Easy (1 c) means instances that are easy for DE and with 1 constraint.}
				{%
					\begin{tabular}{|p{2.2cm}|c|c|c|c|}
					\hline \hline
					Instances	& DE algorithm & ES algorithm & PSO algorithm   \\ \hline						
					DE Easy (1 c) & 24.2K  & 23.6K   & 24.9K  \\ \hline
					ES Easy (1 c) &  24.8K &  24.2K  & 25.4K  \\ \hline
					PSO Easy (1 c) & 26.4K  &  25.4K  & 26.4K  \\ \hline
					DE Easy (2 c) & 25.3K  & 28.1K   & 26.4K  \\ \hline
					ES Easy (2 c) & 24.1K  &  27.2K  & 27.4K  \\ \hline
					PSO Easy (2 c) & 23.5K  &  29.3K  & 271.K  \\ \hline
					DE Easy (3 c) &  27.9K & 31.9K   & 35.5K  \\ \hline
					ES Easy (3 c) &  29.4K &  32.1K  &  28.5K \\ \hline
					PSO Easy (3 c) & 28.1K  & 28.7K   &  29.4K \\ \hline
					DE Easy (4 c) & 34.1K  & 28.9K   & 36.4K  \\ \hline
					ES Easy (4 c) & 35.2K  & 35.3K   &  31.6K \\ \hline
					PSO Easy (4 c) &  31.8K &  29.5K  & 33.2K  \\ \hline
					DE Easy (5 c) &  38.7K & 29.2K   &  37.2K \\ \hline
					ES Easy (5 c) &  35.6K &  28.2K  & 39.5K  \\ \hline
					PSO Easy (5 c) & 36.3K  & 31.5K   & 36.2K  \\ \hline \hline
					DE Hard (1 c) & 129.3K  & 102.7K   & 105.3K  \\ \hline
					ES Hard (1 c) &   104.3K & 121.2K   &  108.2K \\ \hline
					PSO Hard (1 c) & 108.2K  & 104.2K   & 119.8K  \\ \hline
					DE Hard (2 c) &  132.6K & 114.2K   & 114.9K  \\ \hline
					ES Hard (2 c) & 111.2K  & 127.1K   & 112.4K  \\ \hline
					PSO Hard (2 c) & 109.4K  & 112.4K   & 125.3K  \\ \hline
					DE Hard (3 c) & 136.2K  & 116.3K   & 112.4K  \\ \hline
					ES Hard (3 c) & 117.2K  &  132.1K  &  109.9K \\ \hline
					PSO Hard (3 c) & 119.8K  & 119.2K   & 132.6K  \\ \hline
					DE Hard (4 c) & 141.2K  & 119.9K   & 119.6K  \\ \hline
					ES Hard (4 c) & 113.8K  & 131.2K   &  121.9K \\ \hline
					PSO Hard (4 c) & 115.4K  & 121.4K   & 138.9K  \\ \hline
					DE Hard (5 c) & 149.3K  & 129.7K   &  122.9K \\ \hline
					ES Hard (5 c) &  124.4K & 149.6K   &  126.4K \\ \hline
					PSO Hard (5 c) &  123.9K &  124.2K  & 148.3K  \\ \hline							
					\end{tabular}}
				\setlength{\tabcolsep}{5em}
					
					\label{table:perfquadsphere}
				}
			}
			
		\end{table}

	\begin{table}
		\hspace{-1.0cm}
		\resizebox{0.45\columnwidth}{!}{%
			\parbox{.6\linewidth}{
				\centering

				\caption {The comparison of algorithms performance on each other's easy and hard instances based on required FEN for \textbf{Achkley} objective function with \textbf{linear} constraints. DE Easy (1 c) means instances that are easy for DE and with 1 constraint.}
				{%
					\begin{tabular}{|p{2.2cm}|c|c|c|c|}
						
						\hline \hline
						Instances	& DE algorithm & ES algorithm & PSO algorithm   \\ \hline
						DE Easy (1 c) &  39.2K & 37.1K  & 37.4K   \\ \hline
						ES Easy (1 c) &  38.6K &  37.8K  & 38.1K  \\ \hline
						PSO Easy (1 c) & 41.3K  & 39.2K   & 41.7K   \\ \hline
						DE Easy (2 c) & 40.3K  &  41.5K  & 42.6K  \\ \hline
						ES Easy (2 c) &  41.3K & 42.5K   & 41.5K  \\ \hline
						PSO Easy (2 c) & 42.6K  &  41.2K  & 44.7K  \\ \hline
						DE Easy (3 c) &  47.3K & 48.2K   & 46.9K  \\ \hline
						ES Easy (3 c) &  48.9K &  47.3K  & 46.3K  \\ \hline
						PSO Easy (3 c) & 49.2K  &  51.6K  & 50.9K  \\ \hline
						DE Easy (4 c) & 48.9K  & 47.2K   & 49.2K  \\ \hline
						ES Easy (4 c) &  49.2K &  48.2K  & 50.1K  \\ \hline
						PSO Easy (4 c) & 51.2K  & 50.5K   & 52.6K  \\ \hline
						DE Easy (5 c) & 51.3K  & 52.7K   & 51.6K  \\ \hline
						ES Easy (5 c) &  52.1K & 52.6K   & 50.7K  \\ \hline
						PSO Easy (5 c) & 55.3K  & 54.7K   & 52.3K  \\ \hline
						\hline
						DE Hard (1 c) & 107.3K  &  83.2K  & 85.6K  \\ \hline
						ES Hard (1 c) &  82.3K &  105.2K  & 81.6K  \\ \hline
						PSO Hard (1 c) &  85.2K &  83.9K  & 106.3K  \\ \hline
						DE Hard (2 c) & 114.2K  & 88.2K   & 91.5K  \\ \hline
						ES Hard (2 c) &  87.3K &  115.3K  & 88.8K  \\ \hline
						PSO Hard (2 c) &  89.2K &   87.3K &  116.9K \\ \hline
						DE Hard (3 c) &  119.8K &  94.1K  & 93.9K  \\ \hline
						ES Hard (3 c) &  95.2K &  121.6K  &  94.2K \\ \hline
						PSO Hard (3 c) &  93.2K & 95.1K   & 121.5K  \\ \hline
						DE Hard (4 c) & 125.2K  & 99.2K   & 101.4K  \\ \hline
						ES Hard (4 c) &  101.3K &  126.3K  & 98.2K  \\ \hline
						PSO Hard (4 c) &  99.4K &  97.8K  &  127.4K \\ \hline
						DE Hard (5 c) & 132.5K  & 102.2K   & 101.5K  \\ \hline
						ES Hard (5 c) &  101.4K &  134.7K  & 103.5K  \\ \hline
						PSO Hard (5 c) &  103.9K &  102.4K  & 131.4K  \\ \hline
					\end{tabular}}
					\setlength{\tabcolsep}{5em}
					
					\label{table:perflinearackley}
					
				}
			}
			\hfill
			\resizebox{0.45\columnwidth}{!}{%
				\parbox{.6\linewidth}{
					\centering
					\caption {The comparison of algorithms performance on each other's easy and hard instances based on required FEN for \textbf{Ackley} objective function with \textbf{quadratic} constraints. DE Easy (1 c) means instances that are easy for DE and with 1 constraint.}
					{%
						\begin{tabular}{|p{2.2cm}|c|c|c|c|}
							\hline \hline
							Instances	& DE algorithm & ES algorithm & PSO algorithm   \\ \hline						
							DE Easy (1 c) &  38.1K &  39.4K  & 37.2K  \\ \hline
							ES Easy (1 c) &  37.1K & 38.1K   & 39.0K  \\ \hline
							PSO Easy (1 c) &  41.2K & 39.9K   & 40.7K  \\ \hline
							DE Easy (2 c) & 38.9K  &  43.9K  & 41.7K  \\ \hline
							ES Easy (2 c) & 40.1K  & 41.2K   & 43.2K  \\ \hline
							PSO Easy (2 c) &  39.1K &   43.1K &  46.1K \\ \hline
							DE Easy (3 c) & 46.3K  & 47.9K   & 45.1K  \\ \hline
							ES Easy (3 c) & 49.1K  &  48.1K  & 47.2K  \\ \hline
							PSO Easy (3 c) &  42.1K & 45.1K   & 49.1K  \\ \hline
							DE Easy (4 c) &  49.2K &  48.7K  &  48.4K \\ \hline
							ES Easy (4 c) &  49.7K &  49.9K  &   52.9K\\ \hline
							PSO Easy (4 c) & 56.1K  & 55.1K   & 54.1K  \\ \hline
							DE Easy (5 c) &  50.0K & 51.2K   & 52.4K  \\ \hline
							ES Easy (5 c) &  51.3K & 56.2K   & 54.1K  \\ \hline
							PSO Easy (5 c) & 61.2K  & 58.9K   & 59.1K  \\ \hline \hline
							DE Hard (1 c) & 133.4K  &  93.1K  & 94.6K  \\ \hline
							ES Hard (1 c) & 92.1K  &  135.1K  & 94.1K  \\ \hline
							PSO Hard (1 c) & 95.2K  & 94.9K   & 134.2K  \\ \hline
							DE Hard (2 c) & 139.1K  & 98.2K   & 97.1K  \\ \hline
							ES Hard (2 c) &  97.1K &   138.2K &  99.1K \\ \hline
							PSO Hard (2 c) &  109.1K &  107.2K  & 145.2K  \\ \hline
							DE Hard (3 c) &  145.3K &  112.6K  &  109.6K \\ \hline
							ES Hard (3 c) &  111.6K  &  141.2K  & 108.9K  \\ \hline
							PSO Hard (3 c) &  111.2K &  109.2K  & 152.K  \\ \hline
							DE Hard (4 c) & 167.2K  & 132.1K   & 135.1K  \\ \hline
							ES Hard (4 c) & 131.1K   & 167.9K   &  133.9K \\ \hline
							PSO Hard (4 c) & 132.1K  & 133.2K   & 169.1K  \\ \hline
							DE Hard (5 c) & 177.1K   & 143.2K   & 131.3K  \\ \hline
							ES Hard (5 c) &  141.2K &  181.2K  & 144.9K  \\ \hline
							PSO Hard (5 c) & 139.1K  &  142.9K  & 182.1K  \\ \hline							
						\end{tabular}}
						\setlength{\tabcolsep}{5em}
						
						\label{table:perfquadackley}
					}
				}
				
			\end{table}	
\begin{table}
	\hspace{-1.0cm}
	\resizebox{0.45\columnwidth}{!}{%
		\parbox{.6\linewidth}{
			\centering

			\caption {The comparison of algorithms performance on each other's easy and hard instances based on required FEN for \textbf{Rosenbrock} objective function with \textbf{linear} constraints. DE Easy (1 c) means instances that are easy for DE and with 1 constraint.}
			{%
				\begin{tabular}{|p{2.2cm}|c|c|c|c|}
					
					\hline \hline
					Instances	& DE algorithm & ES algorithm & PSO algorithm   \\ \hline
					DE Easy (1 c) &  38.1K & 37.4K   & 39.1K  \\ \hline
					ES Easy (1 c) &  39.2K & 38.9K   & 36.2K  \\ \hline
					PSO Easy (1 c) & 44.6K  &   40.1K & 43.1K  \\ \hline
					DE Easy (2 c) & 41.2K  & 42.4K   & 47.1K  \\ \hline
					ES Easy (2 c) & 40.2K  &  45.2K  & 40.4K  \\ \hline
					PSO Easy (2 c) &  43.9K &   42.6K & 44.8K  \\ \hline
					DE Easy (3 c) &  41.2K & 42.3K   & 45.4K  \\ \hline
					ES Easy (3 c) &  47.7K &   48.1K & 47.4K  \\ \hline
					PSO Easy (3 c) & 48.3K  &  52.4K  & 53.1K  \\ \hline
					DE Easy (4 c) & 44.1K  & 42.7K   & 43.3K  \\ \hline
					ES Easy (4 c) &  48.4K &  49.7K  & 52.6K  \\ \hline
					PSO Easy (4 c) & 53.5K  &  53.1K  &  54.9K \\ \hline
					DE Easy (5 c) & 55.5K  & 53.2K   &  52.1K \\ \hline
					ES Easy (5 c) &  52.8K & 55.4K   &  52.1K \\ \hline
					PSO Easy (5 c) & 58.2K  &   53.4K &  52.8K \\ \hline \hline
					DE Hard (1 c) &  110.2K &  83.7K  & 85.2K  \\ \hline
					ES Hard (1 c) &  81.5K &  107.2K  & 82.1K  \\ \hline
					PSO Hard (1 c) &  86.9K &  85.3K  & 108.2K  \\ \hline
					DE Hard (2 c) & 116.6K  & 89.3K   & 92.1K  \\ \hline
					ES Hard (2 c) &  88.2K &  117.5K  & 87.0K  \\ \hline
					PSO Hard (2 c) &  90.8K &  88.1K  & 114.5K  \\ \hline
					DE Hard (3 c) & 121.2K  & 92.1K   & 92.5K  \\ \hline
					ES Hard (3 c) & 94.1K  &  124.8K  &  93.2K \\ \hline
					PSO Hard (3 c) & 94.7K  &  96.5K  &  125.2K \\ \hline
					DE Hard (4 c) &  126.1K &  101.6K  & 104.2K  \\ \hline
					ES Hard (4 c) & 104.8K  & 127.2K   & 96.1K  \\ \hline
					PSO Hard (4 c) &  100.2K &  92.1K  & 123.7K  \\ \hline
					DE Hard (5 c) &  135.1K &  109.5K  &  106.8K \\ \hline
					ES Hard (5 c) &  105.2K &   136.1K &   105.1K \\ \hline
					PSO Hard (5 c) &  102.1K & 106.8K   & 131.4  \\ \hline
				\end{tabular}}
				\setlength{\tabcolsep}{5em}
				
				\label{table:perflinearrosenbrock}
				
			}
		}
		\hfill
		\resizebox{0.45\columnwidth}{!}{%
			\parbox{.6\linewidth}{
				\centering
				\caption {The comparison of algorithms performance on each other's easy and hard instances based on required FEN for \textbf{Rosenbrock} objective function with \textbf{quadratic} constraints. DE Easy (1 c) means instances that are easy for DE and with 1 constraint.}
				{%
					\begin{tabular}{|p{2.2cm}|c|c|c|c|}
						\hline \hline
						Instances	& DE algorithm & ES algorithm & PSO algorithm   \\ \hline						
						DE Easy (1 c) & 41.2K  & 40.2K   & 38.7K  \\ \hline
						ES Easy (1 c) &  39.2K & 39.3K   & 36.1K  \\ \hline
						PSO Easy (1 c) &  43.1K & 39.6K   & 42.2K  \\ \hline
						DE Easy (2 c) & 39.1K  & 44.4K   & 43.2K  \\ \hline
						ES Easy (2 c) &   42.6K & 44.5K   & 41.8K  \\ \hline
						PSO Easy (2 c) & 41.2K & 45.6K  & 47.2K    \\ \hline
						DE Easy (3 c) &  47.1K &  48.5K  &  49.2K \\ \hline
						ES Easy (3 c) &  46.8K & 49.5K   & 48.8K   \\ \hline
						PSO Easy (3 c) &  46.2K &  42.7K  & 48.4K  \\ \hline
						DE Easy (4 c) &  48.7K &  49.1K  & 51.2K  \\ \hline
						ES Easy (4 c) &  50.2K &  52.4K  & 55.2K  \\ \hline
						PSO Easy (4 c) &  59.2K &  54.5K  &  51.9K \\ \hline
						DE Easy (5 c) &  52.5K &  56.1K  & 55.7K  \\ \hline
						ES Easy (5 c) &  56.0K &  55.7K  & 53.8K  \\ \hline
						PSO Easy (5 c) & 66.3K  & 59.8K   & 60.4K  \\ \hline \hline
						DE Hard (1 c) & 137.2K  & 93.7K   & 92.1K   \\ \hline
						ES Hard (1 c) & 93.7K  & 138.2K   & 99.2K  \\ \hline
						PSO Hard (1 c) & 93.1K  &  96.2K  &  138.0K \\ \hline
						DE Hard (2 c) & 142.7K  & 99.7K   & 98.4K  \\ \hline
						ES Hard (2 c) &  98.8K &  141.6K  & 101.4K  \\ \hline
						PSO Hard (2 c) &  112.7K &  109.5K  & 148.1K  \\ \hline
						DE Hard (3 c) & 148.2K  & 115.3K   & 112.8K  \\ \hline
						ES Hard (3 c) &  114.1K &  144.8K  &  113.2K \\ \hline
						PSO Hard (3 c) & 113.6K  & 113.1K   & 157.3K   \\ \hline
						DE Hard (4 c) &  171.7K &  136.7K  & 133.4K  \\ \hline
						ES Hard (4 c) &  134.7K &  168.2K  &  135.3K \\ \hline
						PSO Hard (4 c) &  134.7K & 139.3K   & 172.6K  \\ \hline
						DE Hard (5 c) &  179.6K &  146.1K  & 144.8K  \\ \hline
						ES Hard (5 c) &  143.8K &   185.1K &  147.4K \\ \hline
						PSO Hard (5 c) & 143.7K  & 141.4K   & 186.4K  \\ \hline							
					\end{tabular}}
					\setlength{\tabcolsep}{5em}
					
					\label{table:perfquadrosenbrock}
				}
			}
			
		\end{table}

\begin{table}
	\hspace{-1.0cm}
	\resizebox{0.45\columnwidth}{!}{%
		\parbox{.6\linewidth}{
			\centering

			\caption {The FEN required for each algorithm to solve  DE/ES/PSO hard instances (Sphere for 1 to 5 \textbf{linear constraints})}
			{%
				\begin{tabular}{|p{2.1cm}|c|c|c|c|}
					
					\hline \hline
					Instances	& DE algorithm & ES algorithm & PSO algorithm   \\ \hline
					DE hard (1 c) & \textbf{86.3K}  &  41.5K  &  43.2K \\ \hline
					ES hard (1 c) & 45.7K & \textbf{84.2K}  & 48.3K    \\ \hline
					PSO hard (1 c)& 37.2K &  41.8K &  \textbf{80.1K}   \\ \hline						
					DE hard (2 c) & \textbf{88.8K}  &  43.9K  &  44.2K\\ \hline
					ES hard (2 c) & 45.9K & \textbf{85.4K}  & 46.3K   \\ \hline
					PSO hard (2 c)& 43.2K &  42.5K &  \textbf{82.9K} \\	\hline
					DE hard (3 c) & \textbf{91.4K}  &  44.6K  &  45.3K \\ \hline
					ES hard (3 c) & 49.2K & \textbf{87.8K}  & 48.1K    \\ \hline
					PSO hard (3 c)& 46.2K &  47.7K &  \textbf{85.5K}   \\ \hline						
					DE hard (4 c) & \textbf{94.2K}  &  47.5K  &  47.8K\\ \hline
					ES hard (4 c) & 51.7K & \textbf{89.1K}  & 50.1K   \\ \hline
					PSO hard (4 c)& 48.7K &  49.9K &  \textbf{87.3K} \\ \hline
					DE hard (5 c) & \textbf{96.2K}  &  48.2K  &  49.5K \\ \hline
					ES hard (5 c) & 52.4K & \textbf{90.4K}  & 53.5K    \\ \hline
					PSO hard (5 c)& 49.6K &  51.4K &  \textbf{91.6K}   \\ \hline							
				\end{tabular}}
				\setlength{\tabcolsep}{5em}
				
				\label{table:fenmultilinsphere}
				
			}
		}
		\hfill
		\resizebox{0.45\columnwidth}{!}{%
			\parbox{.6\linewidth}{
				\centering
				\caption {TThe FEN required for each algorithm to solve  DE/ES/PSO hard instances (Sphere for 1 to 5 \textbf{quadratic constraints})}
				{%
					\begin{tabular}{|p{2.1cm}|c|c|c|c|}
						
						\hline \hline
						Instances	& DE algorithm & ES algorithm & PSO algorithm   \\ \hline
						DE hard (1 c) & \textbf{92.3K}  &  50.2K  &  51.9K \\ \hline
						ES hard (1 c) & 48.8K & \textbf{91.3K}  & 49.3K    \\ \hline
						PSO hard (1 c)& 44.5K &  46.8K &  \textbf{93.1K}   \\ \hline						
						DE hard (2 c) & \textbf{93.5K}  &  52.9K  &  54.2K\\ \hline
						ES hard (2 c) & 50.9K & \textbf{95.9K}  & 51.2K   \\ \hline
						PSO hard (2 c)& 50.2K &  53.2K &  \textbf{96.3K} \\	\hline
						DE hard (3 c) & \textbf{95.9K}  &  54.3K  &  55.3K \\ \hline
						ES hard (3 c) & 53.9K & \textbf{97.4K}  & 52.4K    \\ \hline
						PSO hard (3 c)& 57.3K &  56.3K &  \textbf{98.9K}   \\ \hline						
						DE hard (4 c) & \textbf{98.3K}  &  56.4K  &  57.3K\\ \hline
						ES hard (4 c) & 56.3K & \textbf{102.3K}  & 52.1K   \\ \hline
						PSO hard (4 c)& 59.2K &  58.2K &  \textbf{101.6K} \\ \hline
						DE hard (5 c) & \textbf{102.1K}  &  58.3K  &  59.4K \\ \hline
						ES hard (5 c) & 59.2K & \textbf{103.2K}  & 60.2K    \\ \hline
						PSO hard (5 c)& 62.6K &  63.8K &  \textbf{105.2K}   \\ \hline							
					\end{tabular}}
					\setlength{\tabcolsep}{5em}
					
					\label{table:fenmultiquadsphere}
				}
			}
			
		\end{table}

\begin{table}
	\hspace{-1.0cm}
	\resizebox{0.45\columnwidth}{!}{%
		\parbox{.6\linewidth}{
			\centering

			\caption {The FEN required for each algorithm to solve  DE/ES/PSO hard instances (Ackley for 1 to 5 \textbf{linear constraints})}
			{%
				\begin{tabular}{|p{2.1cm}|c|c|c|c|}
					
					\hline \hline
					Instances	& DE algorithm & ES algorithm & PSO algorithm   \\ \hline
					DE hard (1 c) & \textbf{102.3K}  &  46.1K  &  51.4K \\ \hline
					ES hard (1 c) & 51.2K & \textbf{104.7K}  & 50.2K    \\ \hline
					PSO hard (1 c)& 47.4K &  49.8K &  \textbf{107.4K}   \\ \hline						
					DE hard (2 c) & \textbf{112.1K}  &  56.1K  &  54.1K\\ \hline
					ES hard (2 c) & 53.9K & \textbf{115.9K}  & 48.6K   \\ \hline
					PSO hard (2 c)& 55.5K &  55.3K &  \textbf{117.2K} \\	\hline
					DE hard (3 c) & \textbf{126.1K}  &  63.7K  &  65.2K \\ \hline
					ES hard (3 c) & 59.1K & \textbf{128.3K}  & 58.7K    \\ \hline
					PSO hard (3 c)& 61.7K &  62.8K &  \textbf{134.2K}   \\ \hline						
					DE hard (4 c) & \textbf{124.9K}  &  68.4K  &  63.1K\\ \hline
					ES hard (4 c) & 64.1K & \textbf{129.8K}  & 59.2K   \\ \hline
					PSO hard (4 c)& 67.5K &  69.2K &  \textbf{135.2K} \\ \hline
					DE hard (5 c) & \textbf{138.8K}  &  75.2K  &  74.1K \\ \hline
					ES hard (5 c) & 71.2K & \textbf{137.1K}  & 76.7K    \\ \hline
					PSO hard (5 c)& 73.1K &  74.1K &  \textbf{141.2K}   \\ \hline							
				\end{tabular}}
				\setlength{\tabcolsep}{5em}
				
				\label{table:fenmultilinAckley}
				
			}
		}
		\hfill
		\resizebox{0.45\columnwidth}{!}{%
			\parbox{.6\linewidth}{
				\centering
				\caption {The FEN required for each algorithm to solve  DE/ES/PSO hard instances (Ackley for 1 to 5 \textbf{quadratic constraints})}
				{%
					\begin{tabular}{|p{2.1cm}|c|c|c|c|}
						
						\hline \hline
						Instances	& DE algorithm & ES algorithm & PSO algorithm   \\ \hline
						DE hard (1 c) & \textbf{142.5K}  &  60.1K  &  62.5K \\ \hline
						ES hard (1 c) & 58.5K & \textbf{148.2K}  & 61.4K    \\ \hline
						PSO hard (1 c)& 53.2K &  53.9K &  \textbf{147.7K}   \\ \hline						
						DE hard (2 c) & \textbf{153.3K}  &  58.1K  &  58.1K\\ \hline
						ES hard (2 c) & 59.2K & \textbf{155.5K}  & 59.2K   \\ \hline
						PSO hard (2 c)& 57.8K &  56.3K &  \textbf{157.2K} \\	\hline
						DE hard (3 c) & \textbf{167.3K}  &  65.2K  &  68.1K \\ \hline
						ES hard (3 c) & 63.2K & \textbf{169.2K}  & 69.8K    \\ \hline
						PSO hard (3 c)& 65.7K &  67.9K &  \textbf{167.6K}   \\ \hline						
						DE hard (4 c) & \textbf{174.8K}  &  71.2K  &  75.1K\\ \hline
						ES hard (4 c) & 66.8K & \textbf{169.1K}  & 72.9K   \\ \hline
						PSO hard (4 c)& 69.1K &  68.3K &  \textbf{172.9K} \\ \hline
						DE hard (5 c) & \textbf{179.5K}  &  75.1K  &  76.1K \\ \hline
						ES hard (5 c) & 72.8K & \textbf{174.9K}  & 77.4.2K    \\ \hline
						PSO hard (5 c)& 75.1K &  74.9K &  \textbf{175.9K}   \\ \hline							
					\end{tabular}}
					\setlength{\tabcolsep}{5em}
					
					\label{table:fenmultiquadAckley}
				}
			}
			
		\end{table}

\begin{table}
	\hspace{-1.0cm}
	\resizebox{0.45\columnwidth}{!}{%
		\parbox{.6\linewidth}{
			\centering

			\caption {The FEN required for each algorithm to solve  DE/ES/PSO hard instances (Rosenbrock for 1 to 5 \textbf{linear constraints})}
			{%
				\begin{tabular}{|p{2.1cm}|c|c|c|c|}
					
					\hline \hline
					Instances	& DE algorithm & ES algorithm & PSO algorithm   \\ \hline
					DE hard (1 c) & \textbf{103.1K}  &  48.2K  &  53.9K \\ \hline
					ES hard (1 c) & 53.1K & \textbf{107.2K}  & 52.7K    \\ \hline
					PSO hard (1 c)& 45.7K &  48.1K &  \textbf{109.2K}   \\ \hline						
					DE hard (2 c) & \textbf{115.1K}  &  57.8K  &  55.7K\\ \hline
					ES hard (2 c) & 54.7K & \textbf{113.4K}  & 46.1K   \\ \hline
					PSO hard (2 c)& 54.8K &  54.9K &  \textbf{119.5K} \\	\hline
					DE hard (3 c) & \textbf{124.3K}  &  65.2K  &  62.1K \\ \hline
					ES hard (3 c) & 58.8K & \textbf{127.1K}  & 59.7K    \\ \hline
					PSO hard (3 c)& 62.3K &  65.5K &  \textbf{136.1K}   \\ \hline						
					DE hard (4 c) & \textbf{125.5K}  &  69.1K  &  65.2K\\ \hline
					ES hard (4 c) & 67.5K & \textbf{128.1K}  & 58.7K   \\ \hline
					PSO hard (4 c)& 64.9K &  70.6K &  \textbf{137.1K} \\ \hline
					DE hard (5 c) & \textbf{135.1K}  &  74.1K  &  74.7K \\ \hline
					ES hard (5 c) & 73.7K & \textbf{135.8K}  & 75.1K    \\ \hline
					PSO hard (5 c)& 72.3K &  76.5K &  \textbf{140.9K}   \\ \hline							
				\end{tabular}}
				\setlength{\tabcolsep}{5em}
				
				\label{table:fenmultilinRosen}
				
			}
		}
		\hfill
		\resizebox{0.45\columnwidth}{!}{%
			\parbox{.6\linewidth}{
				\centering
				\caption {The FEN required for each algorithm to solve  DE/ES/PSO hard instances (Rosenbrock for 1 to 5 \textbf{quadratic constraints})}
				{%
					\begin{tabular}{|p{2.1cm}|c|c|c|c|}
						
						\hline \hline
						Instances	& DE algorithm & ES algorithm & PSO algorithm   \\ \hline
						DE hard (1 c) & \textbf{143.1K}  &  61.4K  &  63.7K \\ \hline
						ES hard (1 c) & 59.6K & \textbf{149.7K}  & 62.5K    \\ \hline
						PSO hard (1 c)& 54.3K &  54.2K &  \textbf{143.8K}   \\ \hline						
						DE hard (2 c) & \textbf{155.2K}  &  59.2K  &  59.2K\\ \hline
						ES hard (2 c) & 61.3K & \textbf{154.8K}  & 57.9K   \\ \hline
						PSO hard (2 c)& 59.2K &  57.1K &  \textbf{158.0K} \\	\hline
						DE hard (3 c) & \textbf{168.9K}  &  63.7K  &  66.8K \\ \hline
						ES hard (3 c) & 65.2K & \textbf{170.1K}  & 68.1K    \\ \hline
						PSO hard (3 c)& 63.7K &  68.1K &  \textbf{168.9K}   \\ \hline						
						DE hard (4 c) & \textbf{175.1K}  &  73.8K  &  76.9K\\ \hline
						ES hard (4 c) & 68.2K & \textbf{172.7K}  & 75.2K   \\ \hline
						PSO hard (4 c)& 67.7K &  69.1K &  \textbf{176.2K} \\ \hline
						DE hard (5 c) & \textbf{180.2K}  &  74.2K  &  77.8K \\ \hline
						ES hard (5 c) & 74.2K & \textbf{175.1K}  & 79.2K    \\ \hline
						PSO hard (5 c)& 73.6K &  74.4K &  \textbf{179.4K}   \\ \hline							
					\end{tabular}}
					\setlength{\tabcolsep}{5em}
					
					\label{table:fenmultiquadRosen}
				}
			}
			
		\end{table}
\subsection{Analysis for Linear Constraints }

We run our experiments on Sphere, Ackley and Rosenbrock objective functions. The linear constraints are considered as in Equation \ref{lineareq} with all coefficients $a_{n}$s that are in the range of $[-5,5]$. Also, the problem dimension is set to 30. As it mentioned before, to analyse and discuss some features such as shortest distance, we assume that the optimum is zero ($b\leq 0$). We use three types of problem instances. DE hard denotes problem instances that are hard for DE algorithm but still easy for PSO and ES algorithms. Also, ES hard instances are easy for DE and PSO algorithm in this section. PSO hard means the instances that are hard for PSO but easy for the rest. Each constraint is generated using multi-objective evolver to generate instances that are hard for one algorithm but easy for others. In the following we discuss the features of linear constraints.

\begin{figure*}
	\hbox{\hspace{-18ex}
		\includegraphics[height=0.85\textwidth]{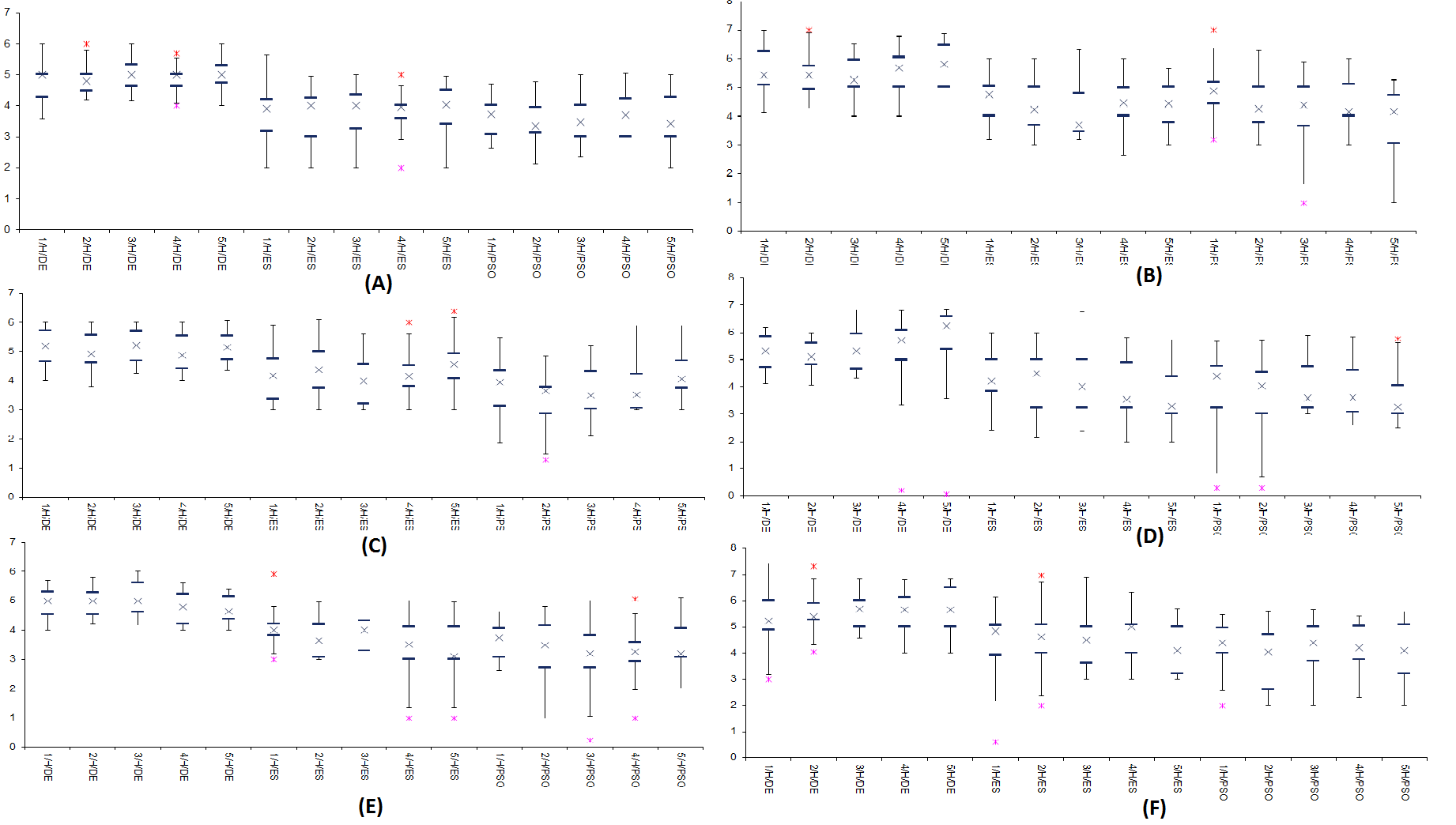}}
	\caption{Box plot for standard deviation of coefficients in linear constraints with objective functions: Sphere (A), Ackley (C) and Rosenbrock (E) and quadratic constraints Sphere (B), Ackley (D) and Rosenbrock (F). Each sub figure includes hard instances (H) with 1 to 5 constraints using algorithms (a/b/c denotes a: number of constraints, b: hard instances and c: hard instances for DE/ES/PSO algorithm).} 
	\label{fig:box_plot_std}
	
\end{figure*}

Figure \ref{fig:box_plot_std} represents some evidence of linear constraint coefficient relationship (standard deviation). It is shown that standard deviation of (1 to 5) linear constraints are higher for DE hard instances than ES and PSO hard ones. This result is similar for all Sphere, Ackley and Rosenbrock objective functions. This means, the instances that are hard for DE algorithm but easy for ES and PSO have higher standard deviation for their constraints coefficients. In other words, this constraint feature has influence on problem difficulty. This improves the prediction ability for algorithm selection framework.  

\begin{figure*}
	\hbox{\hspace{-18ex}
		
		\includegraphics[height=0.9\textwidth]{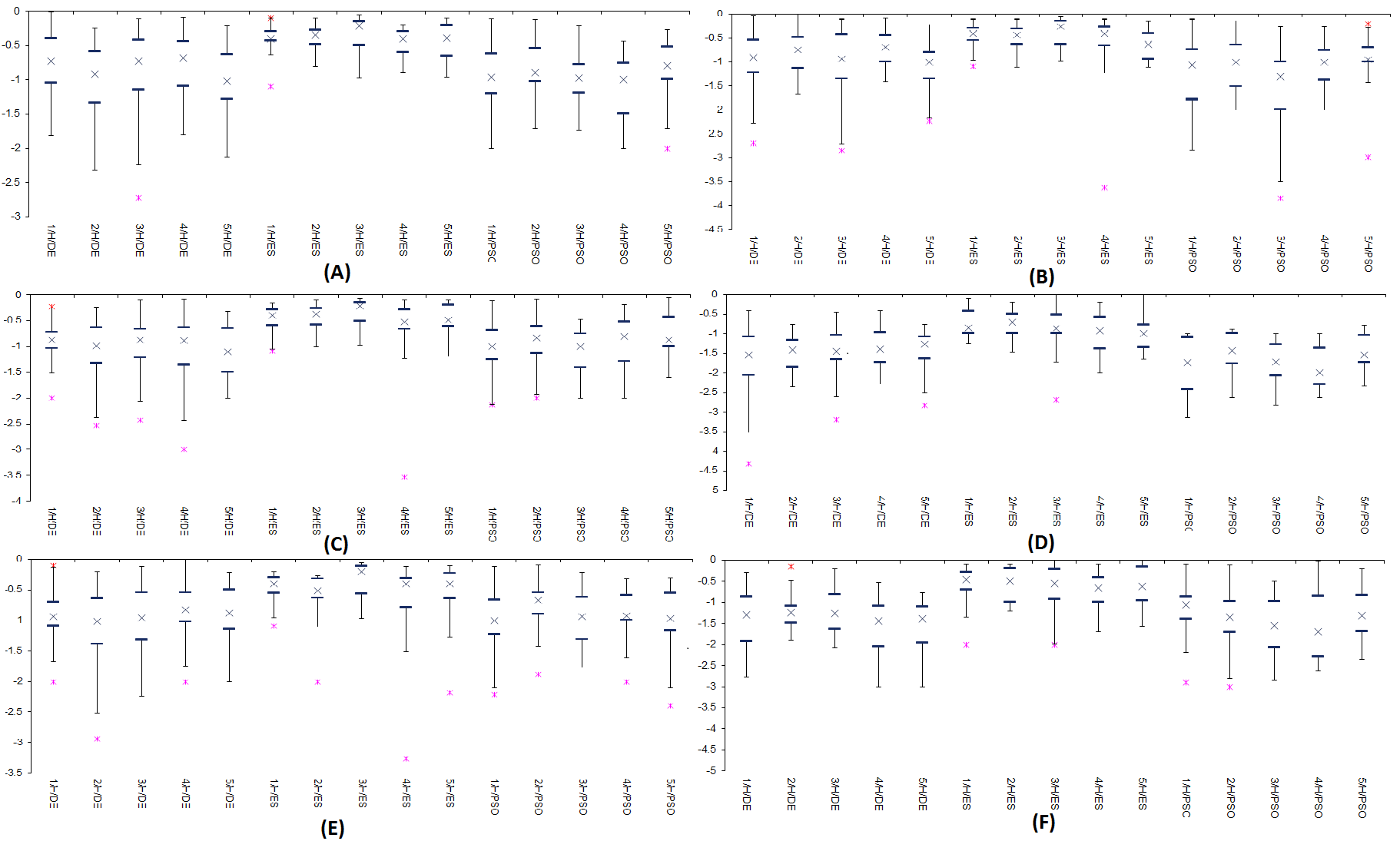}}
	\caption{Box plot for shortest distance feature in linear constraints with objective functions: Sphere (A), Ackley (C) and Rosenbrock (E) and quadratic constraints Sphere (B), Ackley (D) and Rosenbrock (F). Each sub figure includes hard instances (H) with 1 to 5 constraints using algorithms (a/b/c denotes a: number of constraints, b: hard instances and c: hard instances for ES/PSO/DE algorithm).} 
	\label{fig:box_plot_dis}
\end{figure*}

Box plots shown in Figure \ref{fig:box_plot_dis} represent the shortest distance from optimum feature for hard instances. Based on the experiments, hard instances for ES algorithm have higher value (closer to optimum) shortest distance than the other algorithms. It is noteworthy that lower value in Figure \ref{fig:box_plot_dis} means the constraint hyperplane is further from optimum. In other words, the constraints hyperplanes are closer to the optimum in ES hard instances. This relationship holds the pattern for all objective functions in linear constraints. 
We also study the feasibility ratio in vicinity of the optimum. As observed in Table \ref{table:feasibilityratiolin}, hard DE instances have lower feasibility ratio comparing to PSO and ES hard instances. This follows the same pattern for all experimented objective functions. Also increasing the number of constraints decreases the problem optimum-local feasibility for all algorithm problem instances. The angle between linear constraints feature is analysed for linear constraints. As it is observed in Table \ref{table:anglelinear}, ES hard instances have lower angle values for all Sphere, Ackley and Rosenbrock objective functions. This means, instances that are hard for ES have less angle value between their constraint hyperplanes. Interestingly, all objective function that we use in this experiment follow the same relationship.

As it is observed, to compare the instances, DE hard instances have higher linear constraint coefficient standard deviation. It can be translated as DE algorithm has more difficulty to coefficients standard deviation feature than PSO and ES algorithms. Also, the local-optimum feasibility ratio value is higher in ES and PSO hard instances than DE hard ones. This means, ES and PSO algorithms are more effective to problems with higher optimum feasibility ratio feature. The shortest distance and angle features for ES is less than DE and PSO hard instances. Interestingly, this features are similar for all used objective functions. The linear constraint feature based analysis gives us helpful knowledge to implement algorithm selection framework.

\subsection{Analysis for Quadratic Constraints }
In this section, we carry out our experiments on Sphere, Ackley and Rosenbrock objective function with quadratic constraints (see Equation \ref{quadraticeq}) using same setup as previous section. In the following we do feature based analysis of constraints in hard DE, PSO and ES instances (that are easy for the other algorithms).

Figure \ref{fig:box_plot_std} shows some evidence of quadratic constraint coefficients relationship. Based on our experiments, in each constraint, the quadratic coefficient has more ability than linear coefficients to make problem harder to solve. In other words, in Equation \ref{quadraticeq}, $a_{1}$ is more contributing than $a_{2}$ to problem difficulty. As it is shown in the box plots, the standard deviation of 1 to 5 quadratic constraints in DE hard instances are higher comparing the other two algorithm hard instances. In contrast, our results show no systematic relationship between problem difficulty and linear coefficients in each quadratic constraints and quadratic coefficients have more contribution in problem difficulty.

As it is observed in Figure \ref{fig:box_plot_dis}, the shortest distance feature for DE, PSO and ES hard instances are compared. In instances that are hard for ES and easy for the other algorithms, the quadratic constraint hyperplanes are closer to optimum (zero). This applies to all experimented objective functions. 
Also, calculating the angle feature for quadratic constraint does not show any systematic relationship to problem difficulty. The feasibility ratio near the optimum is analysed for DE, ES and PSO hard instances. As it is shown in Table \ref{table:feasibilityratioquad}, the feasibility ratio in DE hard instances are lower than the other algorithms hard instances. All objective functions have the same pattern. Also, the number of constraint has a systematic relationship with feasibility ratio.

Based on the results, to compare COP instances with quadratic constraints, DE hard instances have higher coefficient standard deviation value than the other algorithm hard ones. It is translated as the DE algorithm has more difficulty solving instances with higher standard deviation value for their quadratic constraints than ES and PSO. Also, the quadratic constraints are closer to optimum in ES instances than the other experimented algorithms. In other words, ES algorithm is more influenced by constraint with closer to optimum instances. Moreover, the optimum feasibility ratio in DE instances are lower than PSO and ES.\\

\begin{table}[t]
	\centering
	\caption {The angle feature for Sphere objective function for linear constraints}
	\resizebox{\columnwidth}{1.3 cm}{%
		\begin{tabular}{|p{1.5cm}|p{1cm}|p{1cm}|p{1cm}|p{1cm}|p{1cm}|p{1cm}|p{1cm}|p{1cm}|p{1cm}|p{1cm}|c|}
			
			\hline \hline
			& Cons 1,2 & Cons 1,3 & Cons 1,4 & Cons 1,5 & Cons 2,3 & Cons 2,4 & Cons 2,5 & Cons 3,4 & Cons 3,5 & Cons 4,5 \\ \hline
			
			DE Hard & 74  & 64 &   63  & 58  & 74 & 71  & 68  &   59  & 62  & 86\\ \hline
			ES Hard 	& 33  & 21  &  37  & 24  & 44 & 46  & 39  &   46  & 48  & 51\\ \hline
			PSO Hard & 75  & 63  &   82 & 68  & 71 & 73  & 72  &   69  & 81  & 86 \\ \hline  
			
		\end{tabular}}
		\setlength{\tabcolsep}{5em}
		
		\label{table:anglelinear}
	\end{table}

	\begin{table}[t]
		\caption {Optimum-local feasibility ratio of search space near the optimum for 1,2,3,4 and 5 linear constraint}
		\centering
		\resizebox{0.65\columnwidth}{!}{%
			
			\begin{tabular}{|p{1.5cm}|p{1cm}|p{1cm}|p{1cm}|p{1cm}|p{1cm}|c|}
				
				\hline \hline
				& 1 cons & 2 cons & 3 cons & 4 cons & 5 cons   \\ \hline
				DE Hard & 6\% & 5\%  &  3\%   &  3 \%  & 2\%  \\ \hline
				ES Hard & 16\% &  11\% &    10 \%  &   6\% &  5\% \\ \hline
				PSO Hard & 17\% & 12\%  &   11\%  &   8\% &  5\%  \\ \hline
				
			\end{tabular}}
			\setlength{\tabcolsep}{5em}
			
			\label{table:feasibilityratiolin}
		\end{table}

		\begin{table}[t]
			\caption {Optimum-local feasibility ratio of search space near the optimum for 1,2,3,4 and 5 quadratic constraint}
			\centering
			\resizebox{0.65\columnwidth}{!}{%
				
				\begin{tabular}{|p{1.5cm}|c|c|c|c|c|c|}
					
					\hline \hline
					& 1 cons & 2 cons & 3 cons & 4 cons & 5 cons   \\ \hline
					DE Hard & 4\% & 4\%  &  3\%   &  2 \%  & 2\%  \\ \hline
					ES Hard & 14\% &  10\% &    8 \%  &   7\% &  5\% \\ \hline
					PSO Hard & 15\% & 10\%  &   9\%  &   8\% &  7\%  \\ \hline
					
				\end{tabular}}
				\setlength{\tabcolsep}{5em}
				
				\label{table:feasibilityratioquad}
			\end{table}

\section{Conclusion}
In this paper, we carried out an algorithm performance comparison on each others constrained problem instances. We then analysed the features and characteristics of constraints that make them hard to solve for certain algorithm but easy for the others. It is observed that some constraint features are more contributing to problem difficulty for certain algorithms. In linear constraints, some features such as coefficient relationship, angle, local-optimum feasibility ratio and shortest distance play an important role in problem difficulty to DE and ES algorithms. Considering quadratic instances, angle does not show any relationship to problem difficulty.

By analysing how well one algorithm performs in conditions where other algorithms fail, we can derive its strengths and weaknesses over constrained problems. These results can help us to improve the efficiency of algorithm prediction model.

\section*{Acknowledgements}
Frank Neumann has been supported by ARC grants DP130104395 and DP140103400.

\end{document}